\documentclass[10pt,twocolumn,letterpaper]{article}

\usepackage{cvpr}              

\newcommand{\TODO}[1]{\textbf{\color{red}[TODO: #1]}}
\renewcommand{\TODO}[1]{}

\usepackage{todonotes}
\usepackage{graphicx}
\usepackage[accsupp]{axessibility}
\usepackage{booktabs,makecell,tabularx,siunitx,array,multirow}
\usepackage[table]{xcolor}  

\definecolor{cvprblue}{rgb}{0.21,0.49,0.74}
\usepackage[pagebackref,breaklinks,colorlinks,allcolors=cvprblue]{hyperref}

\def\confName{CVPR}
\def\confYear{2026}

\title{Towards GUI Agents: Vision-Language Diffusion Models for GUI Grounding}

\author{
Shrinidhi Kumbhar$^{1}$\thanks{Work done while at AWS Agentic AI. skumbha4@asu.edu, liahaofu@amazon.com}\quad
Haofu Liao$^{2}$\quad
Srikar Appalaraju$^{2}$\footnotemark[1]\quad
Kunwar Yashraj Singh$^{2}$\\
$^{1}$Arizona State University \quad $^{2}$AWS Agentic AI
}

\begin{document}
\maketitle
\begin{abstract}
Autoregressive (AR) vision–language models (VLMs) have long dominated multimodal understanding, reasoning, and graphical user interface (GUI) grounding. Recently, discrete diffusion vision–language models (DVLMs) have shown strong performance in multimodal reasoning, offering bidirectional attention, parallel token generation, and iterative refinement. However, their potential for GUI grounding remains unexplored. In this work, we evaluate whether discrete DVLMs can serve as a viable alternative to AR models for GUI grounding. We adapt LLaDA-V for single-turn action and bounding-box prediction, framing the task as text generation from multimodal input. To better capture the hierarchical structure of bounding-box geometry, we propose a hybrid masking schedule that combines linear and deterministic masking, improving grounding accuracy by up to 6.1 points in Step Success Rate (SSR) over the GUI-adapted LLaDA-V trained with linear masking. Evaluations on four datasets spanning web, desktop, and mobile interfaces show that the adapted diffusion model with hybrid masking consistently outperforms the linear-masked variant and performs competitively with autoregressive counterparts despite limited pretraining. Systematic ablations reveal that increasing diffusion steps, generation length, and block length improves accuracy but also increases latency, with accuracy plateauing beyond a certain number of diffusion steps. Expanding the training data with diverse GUI domains further reduces latency by about 1.3 seconds and improves grounding accuracy by an average of 20 points across benchmarks. These results demonstrate that discrete DVLMs are a promising modeling framework for GUI grounding and represent an important step toward diffusion-based GUI agents.
\end{abstract}    
\begin{figure*}[t]
  \centering
  \includegraphics[width=\textwidth]{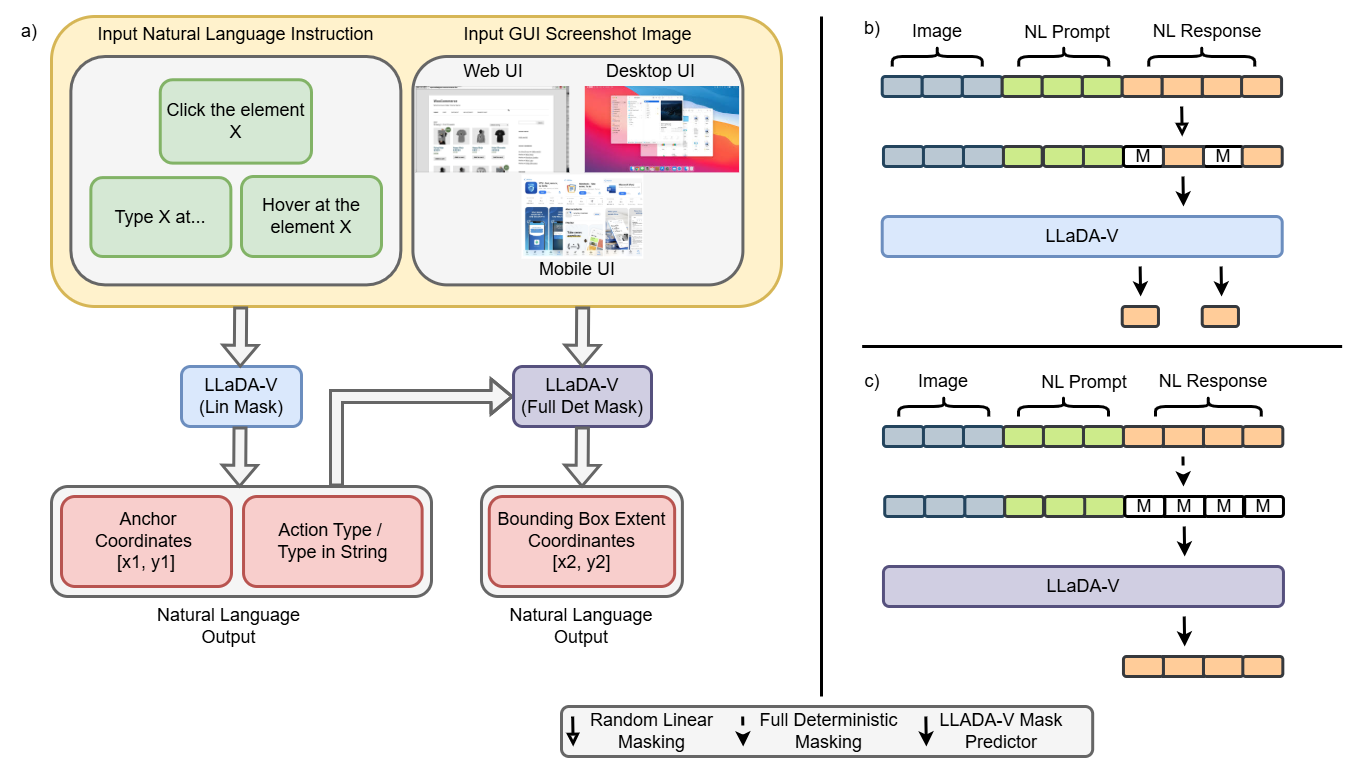}  
  \caption{
  \textbf{Overview of Hybrid Masking Adaptation of LLaDA-V for GUI Grounding.}
  (a) The adapted framework takes a natural-language instruction and a GUI screenshot (either from web, desktop, or mobile interfaces) as input. LLaDA-V trained with the linear masking predicts the action type, optional \texttt{type\_in} text, and anchor coordinates \((x_1, y_1)\). LLaDA-V model trained with the full deterministic masking predicts the remaining bounding-box coordinates \((x_2, y_2)\) conditioned on the image, instruction, and anchor.
  (b) \textbf{Linear Masking Phase:} action and anchor tokens in the response are randomly masked according to the linear masking during the forward corruption process during training for coarse grounding during training.
  (c) \textbf{Deterministic Masking Phase:} all response tokens are fully masked in the forward corruption process during training, and LLaDA-V predicts the bounding-box extent during denoising.
  }
  \label{fig:my_result}
\end{figure*}

\section{Introduction}
Recent advances in multimodal learning have enabled autoregressive (AR) vision-langauge models (VLMs) to jointly reason over images and text, allowing them to describe, interpret, and interact with visual environments in increasingly sophisticated ways. Among the most promising applications of such multimodal understanding and reasoning is the development of GUI agents \citep{hong2024cogagent,nguyen2024gui}, which can perceive digital interfaces and execute actions within them. Within this paradigm, GUI grounding, the ability to locate specific elements in a GUI image from natural language instructions and perform actions on them, serves as a fundamental capability for GUI agents that operate software and automate everyday digital workflows. Each element in the interface is represented by a bounding box \(B = (x_1, y_1, x_2, y_2)\), which defines its spatial region of interaction. For example, given the command ``enter hello in the search bar'' and a webpage screenshot, the model must locate region representing the input field and generate the corresponding action to type the text.

AR VLMs currently dominate GUI grounding research. These models leverage large-scale pretraining on diverse vision–language datasets, including grounding-specific supervision, followed by instruction tuning and post-training refinement. By integrating powerful vision encoders with strong language backbones, they achieve state-of-the-art performance on grounding tasks \cite{bai2025qwen2,team2025kimi,zhang2025phi}. However, AR models inherit architectural properties like sequential decoding and unidirectional attention. These limitations have motivated the exploration of alternative generative paradigms, leading to discrete DVLMs such as LLaDA-V \cite{you2025llada} and MMaDA \cite{yang2025mmada}. These models have shown competitive performance compared to AR VLMs in the mutimodal understanding and reasoning tasks. Yet their potential for GUI grounding is underexplored.
To bridge this gap, we build on the existing discrete DVLM LLaDA-V, adapting it for single-turn action prediction and GUI grounding. LLaDA-V is originally modeled for text generation and reasoning tasks given a multimodal input. It is trained using a discrete diffusion objective that corrupts and denoises tokens through a forward and reverse process with a linear masking schedule. While effective for general multimodal understanding, random token corruption across diffusion steps introduces variations in masked sequences. Such randomness may disrupt the model’s ability to capture consistent geometric dependencies among these coordinates. In GUI grounding, the model must learn geometric relationships between the action type and bounding-box coordinates \(B = (x_1, y_1, x_2, y_2)\), where \((x_1, y_1)\) anchors the action location and \((x_2, y_2)\) defines its spatial extent. 

Motivated by this, we introduce a hybrid masking schedule designed to guide the diffusion process through two complementary phases: one that learns coarse grounding by predicting the action type and anchor coordinates (default linear masking), and another that refines the bounding-box extent conditionally on the anchor (full deterministic masking). This design encourages structured spatial learning while retaining diffusion’s parallel and bidirectional token refinement. Through systematic ablations, we find that increasing diffusion steps, generation length, and block length improves grounding accuracy but also increases latency. Fixing the generation and block lengths to the task-specific output size while varying diffusion steps boosts accuracy up to a threshold before plateauing, beyond which latency rises without further gains. For our setup, using value of 64 for all three parameters offered the best trade-off between accuracy and efficiency. We find, scaling training data with diverse GUI domains accelerates convergence, reducing latency by up to 1.3 seconds and improving grounding accuracy by an average of 20 points across benchmarks. Evaluations on four datasets spanning web, desktop, and mobile interfaces show that our hybrid masking schedule further improves accuracy, achieving up to 6.1 points higher Step Success Rate (SSR) than the GUI-adapted LLaDA-V with default linear masking, and narrowing the gap to AR models from about 25 to under 15 points, despite limited pretraining and no grounding-specific supervision. 

These results demonstrate that discrete diffusion models are a promising step toward vision–language GUI agents and, to the best of our knowledge, represent the first study exploring their use for GUI grounding.

\section{Related Work}
\subsection{GUI Grounding with Autoregressive VLMs}

GUI grounding is a fundamental capability for multimodal GUI agents, enabling them to link perception, language, and action. Several benchmarks like Mind2Web \cite{deng2023mind2web}, ScreenSpot \cite{li2025screenspot}, VisualWebArena \cite{koh2024visualwebarena} have been developed to evaluate the grounding abilities of GUI agents across web, desktop and mobile domains. In this work, we use these benchmarks to evaluate our model on single-step grounding actions.

Most existing approaches to GUI grounding rely on AR VLMs \cite{gao2024enhancing,park2025r,zhu2025turbocharging,zhao2025analysis}, which generate structured action sequences from screenshots and natural language instructions through large-scale pre-training and instruction tuning.
CogAgent \cite{hong2024cogagent} constructs a large GUI–OCR dataset and employs a dual-branch encoder with cross-attention for efficient interface understanding. SeeClick \cite{cheng2024seeclick} enhances VLMs via GUI-specific pretraining for direct element localization, while Ui-Tars \cite{qin2025ui} combines GUI pretraining with task-level reasoning to capture hierarchical dependencies. UGround \cite{gou2024navigating} extends LLaVA \cite{liu2024llavanext} for large-scale synthetic grounding, and OS-Atlas \cite{wu2024atlas}, GUI-R1 \cite{luo2025gui}, and AGUVIS \cite{xu2024aguvis} explore reinforcement tuning, large action modeling, and multi-stage grounding and planning pipelines for generalization.

While discrete DVLMs have recently shown strong performance in multimodal reasoning, their potential as alternatives to autoregressive approaches for GUI grounding remains unexplored. In this work, we explore whether discrete DVLMs can perform GUI grounding by adapting them for single-step action and bounding-box prediction, and introduce hybdrid masking strategy to improve the GUI grounding accuracy.

\subsection{Diffusion Models for Generative and Multimodal Learning}
\subsubsection{Continuous Diffusion Models}

Diffusion models \cite{sohl2015deep,ho2020denoising,song2019generative} have become a leading generative paradigm, achieving strong results in vision \cite{amit2021segdiff,baranchuk2021label,ho2022cascaded,ho2022video,li2022srdiff,meng2021sdedit,saharia2022palette,chen2023diffusiondet} and language \cite{gong2022diffuseq,lin2023text,li2022diffusion,wu2023difformer,strudel2022self,savinov2021step,lovelace2023latent}. These models operate in a continuous latent space, generating data by iteratively denoising Gaussian noise through a learned reverse process. While highly effective for both conditional and unconditional generation, their reliance on continuous representations limits their ability to handle discrete, structured token sequences, such as those required for text generation, given a multimodal input.

\subsubsection{Discrete Diffusion Models}
Discrete diffusion models operate directly on token-level categorical distributions. The forward process progressively masks tokens, and the reverse process reconstructs them through iterative denoising. Prior works differ mainly in their noise schedules and transition mechanisms: D3PM \cite{austin2021structured} uses uniform categorical transitions, DiffusER \cite{reid2022diffuser} models reversible text edits, and others \cite{sahoo2024simple,zheng2023reparameterized,nie2025large} adopt masking-based corruption. However, these methods focus solely on text as input and do not address multimodal text generation, which is essential for interactive vision–language tasks.

Recently, discrete diffusion approach has been extended to multimodal text generation and reasoning. LLaDA-V \cite{you2025llada} integrates visual instruction tuning within masked diffusion, and MMaDA \cite{yang2025mmada} generalizes this framework to a modality-agnostic architecture for text and vision. While these approaches form the foundation of our work, they have not explored multimodal discrete diffusion models for GUI grounding and action generation, a key capability for developing GUI agents. Our work bridges this gap by adapting LLaDA-V a discrete DVLM for GUI grounding and action prediction.

\section{Background}
\subsection{GUI Grounding}

GUI grounding is a fundamental capability for building GUI agents, linking visual perception with language understanding to translate user instructions into executable actions on digital interfaces. 
Given a natural-language instruction \(N\) and a screenshot \(I\) from a web, desktop, or mobile application, the model predicts an action string 
\(a = [a_{\text{type}}, B]\), 
where \(a_{\text{type}} \in \{\texttt{lclick}, \texttt{hover}, \texttt{type\_in}\}\) specifies the action and 
$B = (x_1, y_1, x_2, y_2)$ is the target bounding box, corresponding respectively to clicking, hovering, and typing within the target region.
The model learns a conditional mapping 
$M: (N, I) \rightarrow a$, 
with bounding-box coordinates normalized to $[0, 1000]$ relative to screen size.
 
For instance, \texttt{lclick [42,180,120,250]} clicks within the box, and 
\texttt{type\_in [50,90,200,130] hello} types “hello” in the input field represented by the bounding box. 
A prediction is considered correct only if the predicted action type matches the ground-truth action type and the center of the predicted bounding box lies within the ground-truth bounding box.

We restrict our study to single-turn grounding, where each instruction maps to one atomic action. 
Multi-step planning and dependent actions are left for future work. 
This formulation isolates the model’s core grounding ability, serving as an exploratory step toward diffusion-based GUI agents.

\subsection{LLADA-V}
Our work builds on LLaDA-V~\cite{you2025llada}, a discrete DVLM that combines visual instruction tuning with masked discrete diffusion. 
The architecture comprises a language tower LLaDA \citep{nie2025large}, a vision tower SigLIP-2 \citep{zhai2023sigmoid, tschannen2025siglip}, and a two-layer MLP projector aligning visual embeddings with the language token space. LLaDA-V is trained on multimodal dialogue data involving image--text interactions. 
For a two-turn example, each instance is represented as 
\((v, p^1_0, r^1_0, p^2_0, r^2_0)\), 
where \(v\) is the image embedding, \(p^1_0, p^2_0\) are textual prompts for the first and second turns, and \(r^1_0, r^2_0\) are their corresponding responses. 
The model reconstructs masked tokens using the following discrete diffusion objective:
\begin{equation}
\begin{split}
-\mathbb{E}_{\substack{
v, t, p^1_0, r^1_0, r^1_t,\\
p^2_0, r^2_0, r^2_t}}
\Bigg[
\frac{1}{t}
\sum_{i=1}^{L_{r1}}
\sum_{j=1}^{L_{r2}}
\mathbf{1}\big[r^{1,i}_{t} = [M] \wedge r^{2,j}_{t} = [M]\big] \\
\times \log p_\theta
\big(r^{1,i}_{0}, r^{2,j}_{0} \mid v, p^1_0, r^1_t, p^2_0, r^2_t\big)
\Bigg]
\end{split}
\tag{1}
\end{equation}

where \(r^1_t, r^2_t\) are masked responses at diffusion step \(t\), and \([M]\) denotes the mask token, and i, j are the indexes of the tokens in the two responses.
This objective minimizes the negative log-likelihood of masked tokens, encouraging accurate reconstruction conditioned on both visual and textual context. Empirically, the authors of LLaDA-V found bidirectional attention outperformed causal attention and therefore adopted in LLaDA-V’s attention design.

LLaDA-V is trained through a three-stage process. Stage 1: Language--Image Alignment trains the MLP projector to align visual features with the language embedding space using the LLaVA-NeXT \citep{liu2024llavanext} and the LLaVA-Pretrain dataset \citep{liu2023visual}, while keeping the vision and language towers frozen. 
Stage 2: Visual Instruction Tuning fine-tunes the full model for multimodal instruction following, first on 10M single-image samples from MAmmoTH-VL \citep{guo2025mammoth} to develop image understanding, and then on 2M diverse samples containing single-image, multi-image, and video data to enhance generalization. 
Stage 3: Multimodal Reasoning Enhancement further improves reasoning by training on 900K QA pairs from VisualWebInstruct \citep{jia2025visualwebinstruct}, followed by balanced training that mixes VisualWebInstruct and MAmmoTH-VL and 2M diverse samples with ``/think'' and ``/no\_think'' prompts to balance detailed reasoning and concise answering.

During inference, LLaDA-V generates responses via the reverse process of masked diffusion. 
Given a new prompt, the model conditions on the visual input and prior dialogue context to iteratively denoise a fully masked sequence \(r^1\).  The sequence is refined through successive steps from \(r^t\) to \(r^s\) \((s < t)\) by predicting masked tokens and re-masking a subset based on confidence. Unlike random re-masking, LLaDA-V employs a \textit{low-confidence re-masking strategy}, where uncertain predictions are masked again while confident ones are retained, yielding more stable and accurate generations.

The inference process in LLaDA-V is controlled by three parameters:
\textbf{Diffusion Steps:} A positive integer defined before inference begins. It specifies the number of denoising iterations the model performs to progressively refine its predictions. At each step, the model reconstructs masked tokens with increasing confidence until the sequence is fully unmasked, producing the final response.
\textbf{Generation Length:} A positive integer that defines the length of the output sequence to be generated by the model.
\textbf{Block Length:} The number of sequential blocks used to generate the output sequence. Tokens within each block are generated in parallel, and the block length must be greater than or equal to the generation length. When the two are equal, generation is fully parallel; when the block length is larger, the model generates in a semi-autoregressive manner.

We adopt LLaDA-V as our base model owing to its strong performance in multimodal reasoning and visual instruction following. 
Moreover, at the time of our study, it was the only publicly available discrete DVLM, making it the most suitable and practical foundation for adaptation to GUI grounding.

\section{Method}
\subsection{Adaptation for GUI Grounding}
With LLaDA-V as our base model, we adapt it for single-turn GUI grounding and action prediction. The architecture remains the same as used by the authors, consisting of a language tower (LLaDA), a vision tower (SigLIP-2), and an MLP projector that aligns visual embeddings with the language token space. Below is our adapted mathematical formulation.

Let \(v\) denote the projected image representation (an image of the GUI) and \([M]\) the special mask token. For each training instance comprising one image, one instruction, and one target action string, we denote the tokenized instruction and action sequence as
\[
p^1_0 = [p^{1,i}_0]_{i=1}^{L_{p1}}, \quad r^1_0 = [r^{1,i}_0]_{i=1}^{L_{r1}},
\]
where \(p^1_0\) is the instruction token sequence and \(r^1_0\) represents the tokenized action string (e.g., \texttt{lclick [42,180,120,250]}), and \(L_{p1}\) and \(L_{r1}\) denote the token lengths of the prompt and response, respectively.

We apply a discrete forward diffusion process to corrupt \(r^1_0\) by replacing a proportion of its tokens with \([M]\), obtaining a masked sequence \(r^1_t\). The denoising model \(p_\theta\) is then trained to reconstruct the original tokens conditioned on the masked sequence, instruction, and visual features. The training objective is
\begin{equation}
\begin{split}
L(\theta) = -\,\mathbb{E}_{v, p^1_0, r^1_0, r^1_t, t}
\Bigg[
\frac{1}{t}
\sum_{i=1}^{L_{r1}}
\big[r^{1,i}_{t} = [M] \\
\times\log p_\theta
(r^{1,i}_0 \mid v, p^1_0, r^1_t)
\Bigg]
\end{split}
\tag{2}
\end{equation}
In this setup, the model learns to reconstruct masked action tokens, including action type, coordinates, and text (for \texttt{type\_in}) conditioned on both the visual context and the natural-language instruction.

At test time, we follow the original LLaDA-V inference procedure without modification. The action sequence is initialized as a fully masked response, \(r_1 = [M, \ldots, M]\), and iteratively refined through the reverse masked diffusion process. At each step, the model predicts all masked tokens in parallel, conditioned on the visual and textual inputs. Following LLaDA-V, we apply low-confidence re-masking, retaining high-confidence predictions and re-masking uncertain ones. The process continues until a complete action string is produced, such as \texttt{lclick [42,180,120,250]} or \texttt{type\_in [50,90,200,130] hello}.

\begin{figure*}[t]
    \centering
    \includegraphics[width=\textwidth]{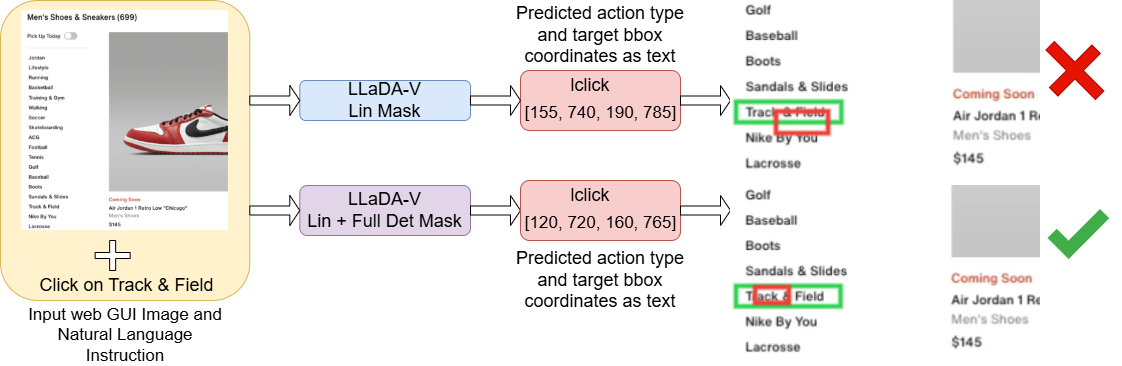}
    \caption{The above figure shows an instance where LLaDA-V 8B trained with linear and full deterministic masking provides a more accurate target bounding box and action prediction compared to the model trained with linear masking. The green bounding box is the ground truth and the red one is the predicted. Both models receive image and natural language instruction and produce action type and target bounding as text, visualised on the GUI image.}
    \label{fig:actual outcome}
\end{figure*}

\subsection{Hybrid Masking Schedule for Structured GUI Grounding}
\label{method}
In LLaDA-V, token masking follows a linear schedule where the probability of masking increases linearly with the diffusion timestep. 
This random corruption treats all tokens action types, brackets, and coordinates as independent, which we hypothesize might overlook the structured relationships inherent in GUI grounding outputs. 
The bounding box \(B = (x_1, y_1, x_2, y_2)\) exhibits a geometric hierarchy: the first pair \((x_1, y_1)\) anchors the action location, while the second pair \((x_2, y_2)\) defines its spatial extent. To model these dependencies, we introduce a hybrid masking schedule with two complementary phases.
\subsubsection{Linear Masking Phase}This phase retains the standard LLaDA-V schedule, allowing the model to learn the action type and anchor coordinates \((x_1, y_1)\) through progressive denoising. 
The masking probability is defined as \(p_{\text{mask}} = (1 - \varepsilon)t + \varepsilon\), 
where \(t \sim \mathcal{U}(0,1)\) is sampled from a uniform distribution and \(\varepsilon\) prevents full masking. Increasing \(t\) increases corruption, exposing the model to varied partial contexts.
\subsubsection{Full Deterministic-Masking Phase.}
In this phase, the model is conditioned on the image \(I\), instruction \(N\), and anchor \((x_1, y_1)\), and trained to predict the remaining coordinates \((x_2, y_2)\) with all target tokens fully masked. 
From a probabilistic standpoint, random linear masking produces limited instances where the extent coordinates are masked while the anchor remains visible an essential condition for learning the conditional relationship between them. 
The deterministic phase enforces this configuration more frequently, effectively strengthening the model’s ability to learn 
\(p_{\theta}(x_2, y_2 \mid a_{\text{type}}, x_1, y_1, I, N)\).

Together, the two phases mirror a coarse-to-fine refinement process first localizing an anchor, then refining it into a complete bounding box improving the model’s ability to capture spatial dependencies and achieve consistent grounding.

\section{Experimental Setup}
\subsection{Datasets}

\begin{table}[t]
\centering
\footnotesize
\setlength{\tabcolsep}{4pt}
\renewcommand{\arraystretch}{1.05}
\caption{\textbf{Training data composition for data scaling experiments.} The dataset spans web, mobile, and desktop domains, totaling 120K samples.}
\begin{tabularx}{\columnwidth}{@{}
  >{\centering\arraybackslash}X  
  >{\centering\arraybackslash}X  
  c c  
  @{}}
\toprule
\textbf{UI Domain} & \textbf{Dataset} & \textbf{Percentage} & \textbf{Samples} \\
\midrule
\multirow{3}{*}{\textbf{Web}} 
 & Mind2Web & 16.67 & 20K \\
 & WebLinX & 16.67 & 20K \\
 & OS-Atlas & 16.67 & 20K \\
 & \textbf{Total} & \textbf{50.00} & \textbf{60K} \\
\midrule
\multirow{2}{*}{\textbf{Mobile}} 
 & OS-Atlas & 16.67 & 20K \\
 & Rico Widget Caption & 16.67 & 20K \\
 & \textbf{Total} & \textbf{33.40} & \textbf{40K} \\
\midrule
\multirow{1}{*}{\textbf{Desktop}} 
 & OS-Atlas & 16.60 & 20K \\
 & \textbf{Total} & \textbf{16.56} & \textbf{20K} \\
\midrule
\rowcolor{gray!10}
\textbf{Overall Total} &  & \textbf{100} & \textbf{120K} \\
\bottomrule
\end{tabularx}
\label{tab:large scale data distribution}
\end{table}

\subsubsection{Training Data}

To validate LLaDA-V on single-step GUI action prediction initially, we trained the model on a 7k subset of the Mind2Web dataset, where each example contains a natural-language instruction, a screenshot of the web UI, and a target action string with bounding-box annotation. To evaluate the impact of visual preprocessing, we use two input variants for training: (1) full, uncropped screenshots, and (2) random crops centered on the target element, where each target element is annotated using the OCR text associated with it.

For the data scaling experiments, we trained on a multi-domain corpus combining four datasets: Mind2Web train split (20K samples), WebLinX (20K samples) \cite{lu2024weblinx}, OS-Atlas (60K samples, 20K each from mobile, web, and desktop domains) \cite{wu2024atlas}, and Rico Widget Caption (20K samples) \cite{deka2017rico,li2020widget}. To handle the large, high-resolution screenshots in Mind2Web, we applied cropping, while for the remaining datasets we used full images. For all datasets, OCR-guided target annotations were used, as this configuration yielded the most reliable grounding performance. The combined dataset, totaling approximately 120K samples, spans mobile, web, and desktop domains, enabling analysis of scaling behavior and multi-domain generalization. The exact data distribution is provided in Table \ref{tab:large scale data distribution}.

\subsubsection{Evaluation Benchmarks}
For evaluation, we used four established GUI grounding benchmarks: Mind2Web (test split), ScreenSpot-Web-Text, ScreenSpot-Web-Icon, and VisualWebArena. M2W, SWT, SWI and VWA are the respective abbreviations we use. These datasets collectively cover a wide range of GUI domains and instruction types.

\subsection{Baselines}
We evaluate Qwen2.5-VL (3B, 7B) \citep{bai2025qwen2} and Phi-3-Vision \citep{team2024phi,zhang2025phi}, representing strong baselines with AR VLMs. For non-autoregressive (NAR) baselines, we use LLaDA-V (8B) in two variants: (1) LLaDA-V (8B) trained on our GUI grounding data under the single-action formulation with its default linear masking schedule; and (2) our proposed hybrid masking variant combining linear and full deterministic masking. All AR models and the LLaDA-V linear and hybrid variants are trained on the same 120K-sample large-scale GUI grounding datasets. For zero shot LLaDA-V refer Appendix B.




\subsection{Inference Parameters for LLaDA-V}
As shown in Tables \ref{tab:small scale sft} and \ref{tab:data quality sft}, we report two step values: Diffusion Steps (Diff. Steps), predefined by the user before inference, and Converged Steps (Conv. Steps), the number of steps after which denoising stops once the model reaches a confident prediction. Our initial study (Table \ref{tab:small scale sft}) shows that setting all inference parameters to 64 yields the best performance, which we adopt for subsequent experiments. In later results (Figure \ref{fig:data_scaling_fullwidth} and Table \ref{tab:AR_VS_NAR}), we report Converged Steps as they reflect the actual number of steps required by LLaDA-V to generate a stable output. For more information, refer to Appendix C.

\subsection{Evaluation Metrics}
\label{SSR}
We evaluate model performance using two metrics: \textbf{Action-Type F1} and \textbf{Step Success Rate (SSR)}. 
Action-Type F1 measures how accurately the model predicts the intended action type among the three possible categories: 
\texttt{lclick}, \texttt{hover}, and \texttt{type\_in}. 
It reflects the balance between precision and recall in classifying actions correctly.

Step Success Rate (SSR) measures whether the predicted bounding box correctly localizes the target element. 
A prediction is considered correct if the center of the predicted bounding box lies within the ground-truth bounding box. 
We use the center point rather than the full area to better represent action precision and robustness, 
as small boundary misalignments in GUI environments often do not affect task success. This criterion provides a reliable and practical measure of GUI grounding accuracy. For equations look Appendix D

\section{Results and Analysis}

\newcolumntype{C}[1]{>{\centering\arraybackslash}m{#1}}
\newcolumntype{Y}{>{\centering\arraybackslash}X}

\begin{table}[t]
\centering
\small
\setlength{\tabcolsep}{4pt}
\renewcommand{\arraystretch}{1.08}
\caption{LLaDA-V 8B fine-tuned only on the Mind2Web training set (7k samples) without cropping and OCR-based target annotation, trained for 10 epochs. The fine-tuned model was evaluated on the Mind2Web test set.}
\begin{tabularx}{\linewidth}{
  Y Y Y Y |  
  Y Y |     
  Y         
}
\toprule
\multicolumn{4}{c|}{\textbf{Inference Parameters}} &
\multicolumn{2}{c|}{\textbf{Accuracy}} &
\textbf{Avg Lat} \\
\midrule  
\textbf{Diff. Steps} & \textbf{Gen Len} & \textbf{Block Len} & \textbf{Conv Steps} &
\textbf{SSR (\%)} & \textbf{F1 (\%)} & \textbf{sec} \\
\midrule
32  & 32  & 32  & 13  & 78.15 & 99.00 & 2.56 \\
64  & 64  & 64  & 25  & 80.67 & 99.00 & 4.84 \\
128 & 128 & 128 & 25  & 80.63 & 99.87 & 5.01 \\
\bottomrule
\end{tabularx}
\label{tab:small scale sft}
\end{table}

\begin{figure*}[t]
    \centering
    \includegraphics[width=\textwidth]{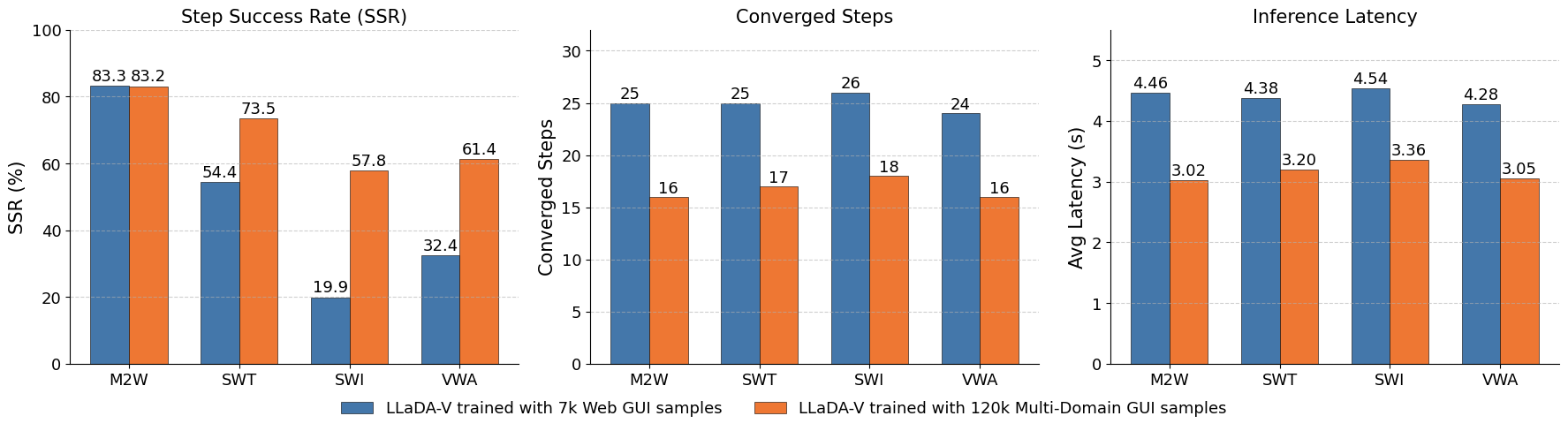}
    \vspace{-2mm}
    \caption{\textbf{GUI Data Scaling behavior of LLaDA-V 8B trained with Linear Masking:} Comparison between LLaDA-V 8B trained on 7k web GUI samples from Mind2Web and 120k mobile, web and desktop GUI samples across four GUI grounding datasets. M2W: Mind2Web, SWT: ScreenSpot-Web-Text, SWI: ScreenSpot-Web-Icon, VWA: Visual Web Arena. Left plot shows Step Sucess Rate (SSR), the center plot shows the number of Converged Steps, and right shows average Inference Latency measured in seconds. Training with large-scale GUI multi-domain data improves SSR, reduces the number of Converged Steps required to produce a highly confident output while reducing Inference Latency, demonstrating better generalization and efficiency across GUI domains.}
    \label{fig:data_scaling_fullwidth}
\end{figure*}

\subsection{Discrete DVLMs Can Perform GUI Grounding}
We first evaluate whether a discrete DVLM model LLaDA-V can perform single step GUI grounding, when trained with the default Linear Masking. As shown in Table \ref{tab:small scale sft}, LLaDA-V 8B, fine-tuned on only 7k Mind2Web samples, achieves an SSR of 80.67\% and an action-type F1 of 99\% on the test set, demonstrating that diffusion-based VLMs show promising capability in predicting both the action type and spatial target from paired image–instruction inputs. To assess generalization, we evaluate the same model on unseen benchmarks. As shown in Fig \ref{fig:data_scaling_fullwidth} despite being trained exclusively on web-based GUIs from Mind2Web, it achieves 54.4\% SSR on ScreenSpot-Web-Text, 19.9\% on ScreenSpot-Web-Icon and 32.4\% on VisualWebArena, whereas the vanilla LLaDA-V model trained on general multimodal data performs near zero. These results indicate that discrete diffusion models generalize beyond their GUI training domain.

The strong performance of the model without grounding-specific pretraining can be attributed to the learning dynamics of diffusion architectures. Its three-stage training process, vision–language alignment, instruction tuning, and multimodal reasoning produces transferable representations that adapt effectively from limited data. The bidirectional attention mechanism enables joint reasoning across action and coordinate tokens, ensuring spatial and semantic consistency, while the iterative denoising process refines localization in a coarse-to-fine manner, where early steps identify approximate regions and later steps use it as a context to enhance precision through low-confidence remasking.

\subsection{Grounding Accuracy and Latency depends on Inference Parameters}
As shown in Table \ref{tab:small scale sft}, increasing diffusion steps, generation length, and block length improves grounding accuracy up to a point but also increases latency. Raising these parameters from 32 to 64 yields an SSR gain of 2.5 points while nearly doubling average latency. Beyond this, accuracy plateaus around 80 percent with minimal variation despite higher inference budgets. These results suggest that larger inference settings do not guarantee better performance and that optimal configurations must be determined empirically. On the Mind2Web test set, 64 diffusion steps and a generation length of 64 tokens provide the best balance between accuracy and latency, while using a block length greater than the generation length increases latency with negligible accuracy gains. Refer Appendix E for more ablations.

\newcolumntype{C}[1]{>{\centering\arraybackslash}m{#1}}
\newcolumntype{Y}{>{\centering\arraybackslash}X}

\begin{table}[t]
\centering
\small
\setlength{\tabcolsep}{4pt}
\renewcommand{\arraystretch}{1.08}
\caption{\textbf{LLaDA-V 8B} fine-tuned only on the \textbf{Mind2Web} training set (7K samples) without and with cropping and OCR-based target annotation, trained for 10 epochs. The highlighted row shows the model trained with random cropping and OCR-based target annotation. Both models were tested on the Mind2Web test set.}
\begin{tabularx}{\linewidth}{
  Y Y Y Y |  
  Y Y |     
  Y         
}
\toprule
\multicolumn{4}{c|}{\textbf{Inference Parameters}} &
\multicolumn{2}{c|}{\textbf{Accuracy}} &
\textbf{Avg Lat} \\
\cmidrule(lr){1-4}\cmidrule(lr){5-6}\cmidrule(lr){7-7}
\textbf{Diff Steps} & \textbf{Gen Len} & \textbf{Block Len} & \textbf{Conv Steps} &
\textbf{SSR (\%)} & \textbf{F1 (\%)} & \textbf{sec} \\
\midrule
64  & 64 & 64 & 25  & 80.67 & 99 & 4.84 \\
\textbf{64} & \textbf{64} & \textbf{64} & \textbf{25} & \textbf{83.31} & \textbf{99} & \textbf{4.46} \\
\bottomrule
\end{tabularx}
\label{tab:data quality sft}
\end{table}
\subsection{Sensitivity to GUI Resolution and Annotation Quality}
We observe that model performance is sensitive to both GUI resolution and annotation quality. In our initial 7k Mind2Web fine-tuning study, the dataset contained screenshots with extremely high image dimensions, which the SigLIP-2 vision encoder in LLaDA-V struggled to process efficiently. Random cropping the images while ensuring the target element remained visible provided more localized and discriminative visual cues, improving the association between instructions and relevant interface regions. Additionally, when ground-truth annotations tightly enclosed icon-level elements rather than the surrounding descriptive text, the model frequently failed to identify the correct target. Refining these annotations to include OCR text regions linked to the target element mitigated this issue. Together, these adjustments yield a +2.68-point SSR improvement and a 0.38 s reduction in latency, underscoring the importance of visual preprocessing and annotation consistency for stable GUI grounding in DVLMs as shown in Table \ref{tab:data quality sft}.

\subsection{Scaling Data Improves Grounding Accuracy, Generalization and Reduces Latency}
Training on a 120K-sample mixture of Mind2Web, WebLinX, OS-Atlas, and RICO datasets significantly improves both accuracy and efficiency. The model achieves an average gain of 17–20 points in SSR and about 5 points in F1, while reducing latency by 1 to 1.5 seconds across benchmarks. Specifically, ScreenSpot-Web-Text improves by +19.1 SSR and +5.2 F1, ScreenSpot-Web-Icon by +37.9 SSR and +8.4 F1, and VisualWebArena by +29 SSR, with all showing reduced latency of roughly 1.2 seconds. On Mind2Web, accuracy remains stable at around 83 percent SSR, with latency decreasing by 1.4 seconds. The model also converges in 8 to 9 fewer diffusion steps, indicating faster denoising and more confident predictions (see Figure \ref{fig:data_scaling_fullwidth}). These results show that training on diverse GUI domains helps the model learn generalizable spatial and semantic priors, improving grounding accuracy and reducing inference time through faster convergence.

\subsection{Hybrid Masking Improves Grounding Accuracy but Increases Latency}
Table \ref{tab:AR_VS_NAR} compares LLaDA-V trained with the default linear masking, our hybrid (linear and full deterministic) masking, and strong AR baselines. The hybrid variant consistently improves grounding accuracy, achieving SSR gains of +1.6 (Mind2Web), +5.3 (ScreenSpot-Web), +1.3 (ScreenSpot-Text), and +6.1 (VisualWebArena) over the linear-only model, while the action-type F1 remains near ceiling. These results support the hypothesis in Section \ref{method} that explicitly conditioning the model to predict the second coordinate pair enhances bounding-box completion and spatial consistency. As illustrated in Figure \ref{fig:actual outcome}, a sample from our evaluation, given the same image and instruction, the model trained with linear masking predicts the correct action type but produces a bounding box that is close to, yet outside, the success criteria defined in the Appendix D. In contrast, the hybrid-masked model generates both the correct action and a precise bounding box that meets the success condition.

Although the hybrid schedule introduces slightly higher latency, this arises from its conditional sequentiality, where the output of the linear phase (action type and anchor coordinates) serves as input for the full deterministic phase that refines the box extent. This dependency adds a mild sequential computation component compared to the default variant. Refer to Appendix F for more results.

\newcolumntype{C}[1]{>{\centering\arraybackslash}m{#1}}
\newcolumntype{Y}{>{\centering\arraybackslash}X}
\newcolumntype{G}{>{\columncolor{gray!15}\centering\arraybackslash}X} 

\begin{table}[t]
\centering
\footnotesize
\setlength{\tabcolsep}{3.0pt}
\renewcommand{\arraystretch}{1.4}
\caption{\textbf{AR vs. NAR GUI grounding comparison.} The table reports GUI grounding performance on four benchmarks across the listed evaluation metrics. The rightmost column (in grey) highlights our proposed LLaDA-V variant trained with the hybrid linear and deterministic masking schedule.}
\begin{tabularx}{\columnwidth}{@{}
  >{\centering\arraybackslash}l
  >{\centering\arraybackslash}l
  >{\centering\arraybackslash}X
  >{\centering\arraybackslash}X
  >{\centering\arraybackslash}X
  >{\centering\arraybackslash}X
  | G @{}}
\toprule
\textbf{Dataset} & \textbf{Metric} &
\textbf{Phi (3B)} &
\textbf{Qwen 2.5 VL (3B)} &
\textbf{Qwen 2.5 VL (7B)} &
\textbf{LLaDA-V 8B (Lin)} &
\textbf{LLaDA-V 8B (Ours)} \\
\midrule
M2W & Conv Steps & 21.00 & 21.00 & 21.00 & 16.00 & 23.00 \\
    & SSR (\%) & 56.80 & 79.30 & 81.90 & 82.40 & 83.90 \\
    & F1 (\%) & 94.40 & 99.60 & 99.90 & 98.50 & 100.00 \\
    & Lat. (s) & -- & -- & 1.10 & 3.02 & 5.44 \\
\midrule
SWI & Conv Steps & 23.00 & 23.00 & 23.00 & 18.00 & 23.00 \\
    & SSR (\%) & 62.60 & 79.10 & 85.40 & 57.80 & 63.10 \\
    & F1 (\%) & 100.00 & 100.00 & 100.00 & 99.50 & 99.60 \\
    & Lat. (s) & -- & -- & 1.10 & 3.36 & 6.50 \\
\midrule
SWT & Conv Steps & 22.00 & 22.00 & 22.00 & 17.00 & 23.00 \\
    & SSR (\%) & 77.00 & 83.00 & 83.00 & 73.50 & 74.80 \\
    & F1 (\%) & 100.00 & 100.00 & 100.00 & 99.10 & 99.60 \\
    & Lat. (s) & -- & -- & 1.10 & 3.20 & 4.20 \\
\midrule
VWA & Conv Steps & 23.00 & 23.00 & 23.00 & 16.50 & 23.00 \\
    & SSR (\%) & 68.50 & 88.90 & 87.20 & 61.40 & 67.50 \\
    & F1 (\%) & 99.90 & 99.80 & 99.90 & 99.40 & 99.90 \\
    & Lat. (s) & -- & -- & 1.10 & 3.05 & 5.49 \\
\bottomrule
\end{tabularx}
\label{tab:AR_VS_NAR}
\end{table}

\section{Conclusion}
This work presents the first systematic study of discrete DVLMs for GUI grounding, a capability essential for multimodal GUI agents. We adapt LLaDA-V for single-turn grounding and action prediction and analyze its behavior across inference dynamics, data scaling, visual preprocessing, and masking strategies. Our results show that diffusion-based models can accurately predict action types and bounding boxes, generalize across domains, and achieve competitive grounding performance despite limited pretraining compared to autoregressive counterparts. We also propose a hybrid masking schedule that captures the structured dependency between anchor and extent coordinates, yielding consistent gains in Step Success Rate (SSR) across benchmarks. Overall, our findings highlight DVLMs as a viable and scalable alternative to sequential autoregressive approaches, establishing a foundation for future research on diffusion-driven GUI agents.


{
    \small
    \bibliographystyle{ieeenat_fullname}
    \bibliography{main}
}

\clearpage
\setcounter{page}{1}
\maketitlesupplementary

\section{Discussion and Limitations}
\label{sec:rationale}

While our work proposes an improved grounding technique for diffusion vision–language models, it remains an early step toward bridging the gap with AR approaches. Although diffusion models demonstrate promising structured grounding capabilities, they still lag behind AR models in latency and accuracy, possibly due to the latter’s extensive grounding-specific pretraining and optimized decoding strategies. Our current setup focuses on single-step action prediction, where the output length is short and the latency difference remains modest. However, extending to multi-step action prediction, which involves longer output sequences, will likely present different outcomes. Future work should explore grounding-specific pretraining, efficient diffusion decoding strategies, and architectural refinements to scale diffusion-based models toward multi-action GUI grounding. 

\section{Appendix B }
\label{APPENDIX B}
We evaluated the performance of LLaDA-V with zero-shot prompting. We observed that the model had near-zero performance for the bounding box and action prediction as shown in Table \ref{tab:zero shot}

\newcolumntype{C}[1]{>{\centering\arraybackslash}m{#1}}
\newcolumntype{Y}{>{\centering\arraybackslash}X}

\begin{table}[t]
\centering
\small
\setlength{\tabcolsep}{4pt}
\renewcommand{\arraystretch}{1.08}
\caption{Zero Shot performance of LLaDA-V 8B on the Mind2Web test set}
\begin{tabularx}{\linewidth}{
  Y Y Y |  
  Y Y     
}
\toprule
\multicolumn{3}{c|}{\textbf{Inference Parameters}} &
\multicolumn{2}{c}{\textbf{Accuracy}} \\
\midrule
\textbf{Diff. Steps} & \textbf{Gen Len} & \textbf{Block Len} &
\textbf{SSR (\%)} & \textbf{F1 (\%)} \\
\midrule
32  & 32  & 32  & 0.00 & 0.10 \\
64  & 64  & 64  & 0.00 & 0.12 \\
128 & 128 & 128 & 0.00 & 0.10 \\
256 & 64  & 64  & 0.00 & 0.10 \\
\bottomrule
\end{tabularx}
\label{tab:zero shot}
\end{table}

\section{Appendix C}
\label{APPENDIX C}
We trained LLaDA-V on 7k samples of the Mind2Web train split. For this study, we didn't use OCR-based annotation and cropped images. The sole purpose of this study was to verify if LLaDA-V has potential for GUI grounding and action prediction. We tested the model with different inference parameters and observed that keeping a value of 64 for all inference parameters gives us the best results. The reason is that the model needs a sequence length of 64 as it represents the max length of the target sequence in the test set. The results are shown in  Table \ref{tab:small scale sft full inf perf}

\newcolumntype{C}[1]{>{\centering\arraybackslash}m{#1}}
\newcolumntype{Y}{>{\centering\arraybackslash}X}

\begin{table}[t]
\centering
\small
\setlength{\tabcolsep}{4pt}
\renewcommand{\arraystretch}{1.08}
\caption{LLaDA-V 8B fine-tuned only on the Mind2Web training set (7k samples) without cropping and OCR based annotation. The table shows the effect of inference parameters on Avg Latency and Accuracy.}
\begin{tabularx}{\linewidth}{
  Y Y Y Y |  
  Y Y |     
  Y         
}
\toprule
\multicolumn{4}{c|}{\textbf{Inference Parameters}} &
\multicolumn{2}{c|}{\textbf{Accuracy}} &
\textbf{Avg Lat} \\
\midrule  
\textbf{Diff. Steps} & \textbf{Gen Len} & \textbf{Block Len} & \textbf{Conv Steps} &
\textbf{SSR (\%)} & \textbf{F1 (\%)} & \textbf{sec} \\
\midrule
32  & 32  & 32  & 13  & 78.15 & 99.00 & 2.56 \\
64  & 64  & 64  & 25  & 80.67 & 99.00 & 4.84 \\
128 & 128 & 128 & 25  & 80.63 & 99.87 & 5.01 \\
256 & 64 & 64 & 25  & 80.69 & 99.87 & 4.84 \\
\bottomrule
\end{tabularx}
\label{tab:small scale sft full inf perf}
\end{table}

\section{Appendix D}
\label{SSR}

We evaluate model performance using two metrics: \textbf{Action-Type F1} and \textbf{Step Success Rate (SSR)}.

\paragraph{Action-Type F1.}
The Action-Type F1 measures how accurately the model predicts the intended action type among the three possible categories (\texttt{lclick}, \texttt{hover}, and \texttt{type\_in}). 
It is defined in terms of precision and recall as:
\[
\text{Precision} = \frac{TP}{TP + FP}, \quad 
\text{Recall} = \frac{TP}{TP + FN},
\]
\[
\text{F1} = \frac{2 \times \text{Precision} \times \text{Recall}}
{\text{Precision} + \text{Recall}},
\]
where \(TP\), \(FP\), and \(FN\) represent true positives, false positives, and false negatives, respectively. 
The macro-averaged F1 is computed across all three action types.

\paragraph{Step Success Rate (SSR).}
The Step Success Rate quantifies how well the predicted bounding box localizes the target element. 
Let the predicted box be \(B_p = (x^p_1, y^p_1, x^p_2, y^p_2)\) and the ground truth box be \(B_g = (x^g_1, y^g_1, x^g_2, y^g_2)\). 
The center of the predicted box is computed as:
\[
c_p = \left(\frac{x^p_1 + x^p_2}{2}, \frac{y^p_1 + y^p_2}{2}\right).
\]
A prediction is considered correct if \(c_p \in B_g\). 
The SSR is then defined as:
\[
\text{SSR} = \frac{1}{N} \sum_{i=1}^{N} 
\mathbf{1}\big[c^{(i)}_p \in B^{(i)}_g\big],
\]
where \(\mathbf{1}[\cdot]\) is the indicator function and \(N\) is the total number of evaluated instances.

\newcolumntype{C}[1]{>{\centering\arraybackslash}m{#1}}
\newcolumntype{Y}{>{\centering\arraybackslash}X}
\newcolumntype{G}{>{\columncolor{gray!15}\centering\arraybackslash}X} 
\newcolumntype{H}{>{\centering\arraybackslash}X}                      

\begin{table}[t]
\centering
\footnotesize
\setlength{\tabcolsep}{3pt}
\renewcommand{\arraystretch}{1.3}
\caption{\textbf{Accuracy–Latency Trade-Off with Hybrid Masking.} The table compares LLaDA-V trained with default linear masking (third column) and our hybrid masking (fourth and fifth columns) across four benchmarks. The hybrid model achieves higher SSR but with slightly higher latency, while reducing diffusion steps lowers latency with minimal loss in accuracy.}
\begin{tabularx}{\columnwidth}{@{}
  >{\centering\arraybackslash}l
  >{\centering\arraybackslash}l
  >{\centering\arraybackslash}X
  | G H @{}}
\toprule
\textbf{Dataset} & \textbf{Metric} &
\textbf{LLaDA-V 8B (Lin)} &
\multicolumn{2}{c}{\textbf{LLaDA-V 8B (Ours)}} \\

\midrule
M2W & Conv Steps & 16.00 & 23.00 & 11.00 \\
    & SSR (\%) & 82.40 & 83.90 & 81.00 \\
    & F1 (\%) & 98.50 & 100.00 & 96.66 \\
    & Lat. (s) & 3.02 & 5.44 & 2.74 \\
\midrule
SWI & Conv Steps & 18.00 & 23.00 & 11.00 \\
    & SSR (\%) & 57.80 & 63.10 & 59.60 \\
    & F1 (\%) & 99.50 & 99.60 & 98.20 \\
    & Lat. (s) & 3.36 & 6.50 & 2.93 \\
\midrule
SWT & Conv Steps & 17.00 & 23.00 & 15.00 \\
    & SSR (\%) & 73.50 & 74.80 & 70.00 \\
    & F1 (\%) & 99.10 & 99.60 & 90.00 \\
    & Lat. (s) & 3.20 & 4.20 & 3.00 \\
\midrule
VWA & Conv Steps & 16.50 & 23.00 & 11.00 \\
    & SSR (\%) & 61.40 & 67.50 & 59.20 \\
    & F1 (\%) & 99.40 & 99.90 & 99.85 \\
    & Lat. (s) & 3.05 & 5.49 & 2.87 \\
\bottomrule
\end{tabularx}
\label{tab:AR_VS_NAR_app}
\end{table}

\section{Appendix E}
We trained LLaDA-V on 7k samples from the Mind2Web training split using OCR-based annotations and cropped images Table \ref{tab:latency_full_perf_app}, following the inference configuration described in Section \ref{APPENDIX C}. To study the effect of inference parameters, we varied the number of diffusion steps, generation length, and block length to observe their impact on grounding accuracy and latency. We found that increasing these parameters improves accuracy but also increases latency, with accuracy eventually plateauing or slightly declining when all parameters are raised simultaneously. Interestingly, the converged steps remain lower than the total diffusion steps around 25 on average even when diffusion steps are increased, indicating that LLaDA-V typically reaches confident predictions early in the denoising process.

\newcolumntype{C}[1]{>{\centering\arraybackslash}m{#1}}
\newcolumntype{Y}{>{\centering\arraybackslash}X}

\begin{table*}[t]
\centering
\small
\setlength{\tabcolsep}{5pt}
\renewcommand{\arraystretch}{1.1}
\caption{\textbf{LLaDA-V 8B inference performance across different inference settings.} 
The model was fine-tuned on the Mind2Web training set (7k samples) with cropping and OCR-based target annotation, trained for 10 epochs.}
\begin{tabularx}{\textwidth}{
  Y Y Y Y |  
  Y Y |     
  Y Y Y     
}
\toprule
\multicolumn{4}{c|}{\textbf{Inference Parameters}} &
\multicolumn{2}{c|}{\textbf{Accuracy}} &
\multicolumn{3}{c}{\textbf{Latency (seconds)}} \\
\midrule
\textbf{Diff. Steps} & \textbf{Gen Len} & \textbf{Block Len} & \textbf{Conv Steps} &
\textbf{SSR (\%)} & \textbf{F1 (\%)} &
\textbf{Lowest} & \textbf{Highest} & \textbf{Avg} \\
\midrule
8   & 64  & 32  & 4   & 71.65 & 99.00 & 0.99 & 2.14 & 1.27 \\
16  & 64  & 32  & 8   & 74.82 & 99.00 & 1.14 & 2.76 & 1.47 \\
32  & 32  & 32  & 13  & 82.50 & 99.00 & 2.24 & 3.43 & 2.56 \\
64  & 64  & 64  & 25  & 83.31 & 99.00 & 3.72 & 10.35 & 4.46 \\
128 & 128 & 128 & 25  & 82.92 & 99.00 & 4.01 & 28.13 & 5.02 \\
256 & 64  & 64  & 25  & 80.61 & 99.00 & 3.83 & 11.03 & 4.81 \\
\bottomrule
\end{tabularx}
\label{tab:latency_full_perf_app}
\end{table*}

\section{Appendix F}
The following subsections present the results obtained using cropped images with OCR-based annotations and the hybrid masking approach, which combines linear and deterministic full masking, to analyze their impact on grounding accuracy and latency.

\subsection{Sensitivity to GUI Resolution and Annotation Quality}
\begin{figure*}[t]
    \centering
    \includegraphics[width=\textwidth]{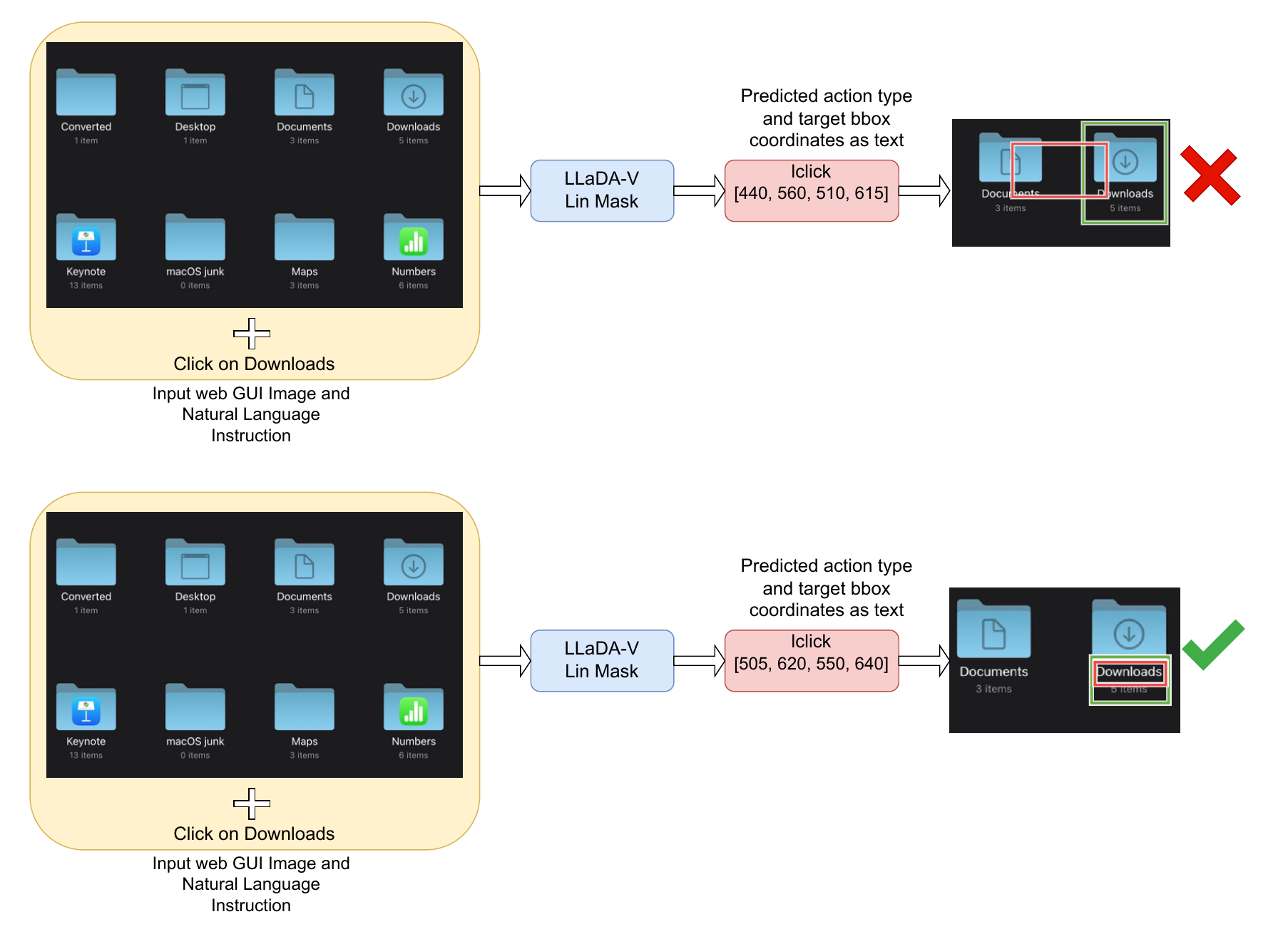}
\caption{\textbf{Effect of data annotation quality on grounding accuracy.} The figure compares predictions from LLaDA-V trained with default linear masking using the Mind2Web train split with and without cropped images and OCR-based annotations. The top example, trained without OCR text-based annotations and cropping, produces an inaccurate bounding box due to inconsistent icon-level targets and high-resolution inputs. The bottom example, trained with cropped images and OCR-guided text annotations, provides more stable supervision, allowing the model to correctly localize the target element. The green bounding box is the ground truth and the red one is the prediction.}
    \label{fig:effect of data_annot_appendix}
\end{figure*}

As shown in Figure \ref{fig:effect of data_annot_appendix}, we compare predictions from the LLaDA-V model trained with default linear masking, using the Mind2Web train set with and without cropped images and OCR-based annotations. The top example shows the model trained without cropping or OCR annotations. In this case, the model struggles with high-resolution inputs and inconsistent element annotations based on icons of varying sizes, leading to inaccurate grounding of target elements. In contrast, the bottom example shows the model trained with cropped images and OCR-based annotations. Cropping reduces visual complexity, and using text-associated annotations provides more consistent supervision, enabling the model to accurately locate the target element.

\subsection{Effect of Hybrid Masking combining Linear and Full Deterministic Masking}
\begin{figure*}[t]
    \centering
    \includegraphics[width=\textwidth]{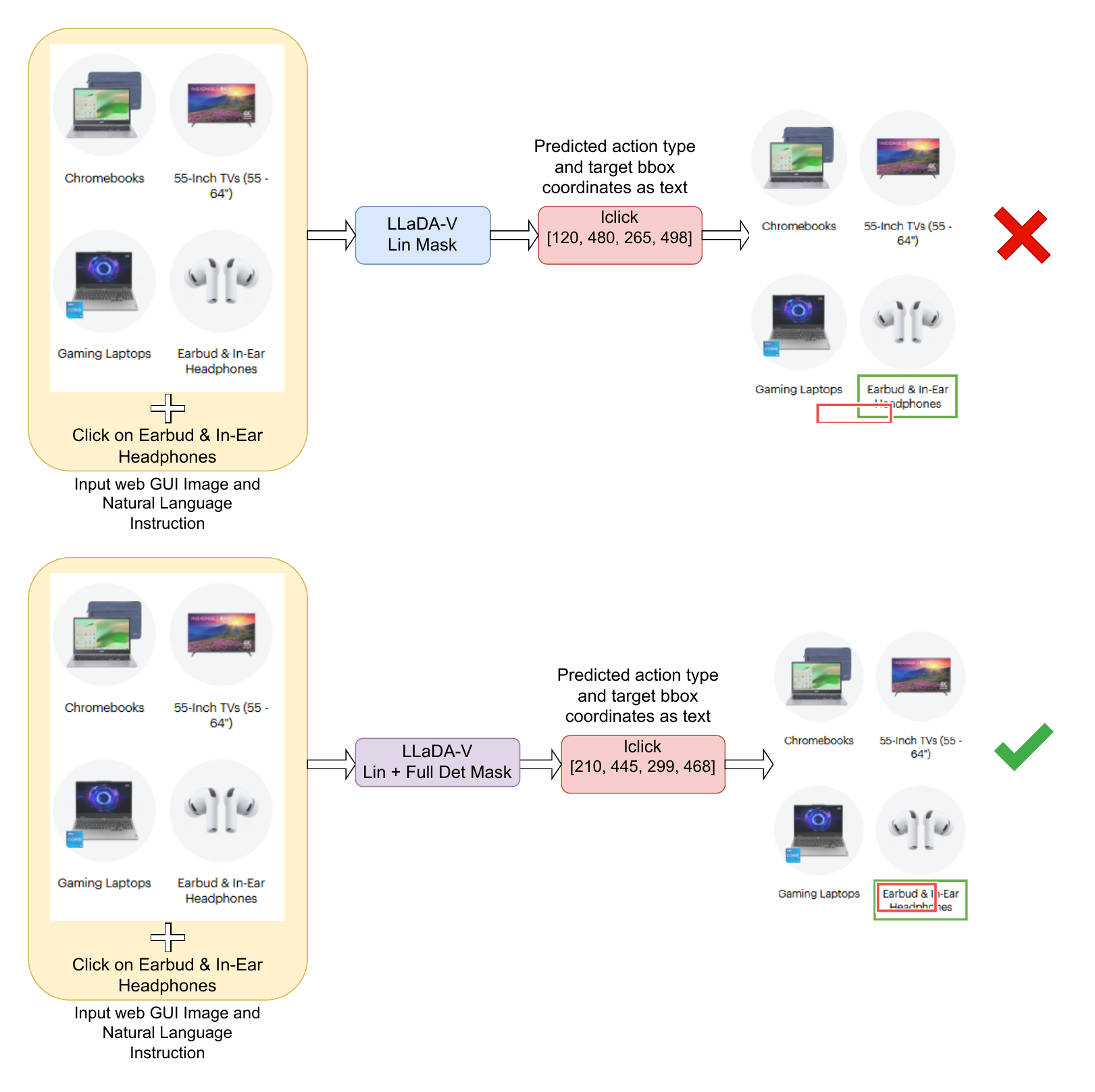}
\caption{\textbf{Effect of hybrid masking on bounding-box accuracy.} The figure compares predictions from LLaDA-V trained with default linear masking (top) and with the proposed hybrid masking that combines linear and deterministic full masking (bottom). The linear-masked model correctly predicts the action type but generates an inaccurate bounding box, missing the target region. In contrast, the hybrid-masked model, guided by conditional refinement between anchor and extent coordinates, produces a precise bounding box that accurately localizes the target element. The green bounding box is the ground truth and the red one is the prediction.}
    \label{fig: hybrid masking appendix}
\end{figure*}

As shown in Figure \ref{fig: hybrid masking appendix}, we compare the predictions made by LLaDA-V trained with the default linear masking and our proposed hybrid masking schedule, which combines linear and deterministic full masking(used cropped images and OCR-based annotation for training both variants). The top example shows the model trained with only linear masking, which correctly predicts the action type but produces a slightly misaligned bounding box that does not meet the success criteria. In contrast, the hybrid-masked model (bottom) generates both the correct action and a precisely localized bounding box that satisfies the success condition.

This improvement can be attributed to the structured conditioning introduced by the hybrid schedule. The linear phase enables coarse grounding by predicting the action type and anchor coordinates, while the deterministic phase explicitly conditions the model to refine the bounding-box extent based on the anchor. This targeted supervision helps the model capture geometric dependencies between coordinates, leading to more accurate and spatially consistent predictions across complex GUI layouts.
\subsection{Latency Accuracy Tradeoff with Hybrid Masking}
Table \ref{tab:AR_VS_NAR_app} compares LLaDA-V trained with default linear masking (third column) and our proposed hybrid masking (fourth and fifth columns) across four benchmarks. The hybrid model achieves higher grounding accuracy, with notable SSR gains across all datasets, but also incurs increased latency due to its conditional sequentiality, where the deterministic phase refines predictions based on the linear phase outputs. When diffusion steps are reduced (fifth column), latency decreases substantially with only a slight drop in accuracy, illustrating a clear trade-off between precision and efficiency in diffusion-based grounding.

\end{document}